\relax
\documentclass[letterpaper]{article} 
\usepackage{aaai22}  
\usepackage{times}  
\usepackage{helvet}  
\usepackage{courier}  
\usepackage[hyphens]{url}  
\usepackage{graphicx} 
\usepackage{natbib}  
\usepackage{caption} 
\DeclareCaptionStyle{ruled}{labelfont=normalfont,labelsep=colon,strut=off} 
\usepackage{algorithm}
\usepackage{algorithmic}
\usepackage{amsmath}
\usepackage{amssymb}

\usepackage{multirow}
\usepackage{adjustbox}
\usepackage{enumitem}

\usepackage{newfloat}
\usepackage{listings}
\floatstyle{ruled}
\newfloat{listing}{tb}{lst}{}
\floatname{listing}{Listing}
\title{Towards Adversarially Robust Deep Metric Learning}
\author{
    Xiaopeng Ke\\
}
\affiliations{

    mezereonxp@gmail.com

}

\usepackage{bibentry}

\begin{document}
\maketitle

\begin{abstract}
    Deep Metric Learning (DML)  has shown remarkable successes in many domains by taking advantage of powerful deep neural networks. Deep neural networks are prone to adversarial attacks and could be easily fooled by adversarial examples. The current progress on this robustness issue is mainly about deep classification models but pays little attention to DML models. Existing works fail to thoroughly inspect the robustness of DML and neglect an important DML scenario, the clustering-based inference. 
    
    In this work, we first point out the robustness issue of DML models in clustering-based inference scenarios. We find that, for the clustering-based inference, existing defenses designed DML are unable to be reused and the adaptions of defenses designed for deep classification models cannot achieve satisfactory robustness performance. To alleviate the hazard of adversarial examples, we propose a new defense, the Ensemble Adversarial Training (EAT), which exploits ensemble learning and adversarial training. EAT promotes the diversity of the ensemble, encouraging each model in the ensemble to have different robustness features, and employs a self-transferring mechanism to make full use of the robustness statistics of the whole ensemble in the update of every single model. We evaluate the EAT method on three widely-used datasets with two popular model architectures. The results show that the proposed EAT method greatly outperforms the adaptions of defenses designed for deep classification models.
\end{abstract}

\section{Introduction}

Deep Metric Learning (DML) aims to learn a distance metric over objects using deep neural networks. Powered by the great success of deep learning techniques, DML has demonstrated remarkable performance in many critical domains, including face verification \cite{hu2014discriminative, schroff2015facenet, lu2017discriminative, liu2017adaptive}, pedestrian re-identification \cite{shi2015constrained,xiao2017joint,wojke2018deep, huang2021full}, representation learning \cite{bengio2013representation,wang2015deep,qiao2019transductive,he2020momentum}, few-shot learning \cite{ravi2016optimization, sung2018learning, kim2019deep, wang2020generalizing}, etc.

\begin{figure}[h]
\begin{center}
\includegraphics[width=\linewidth]{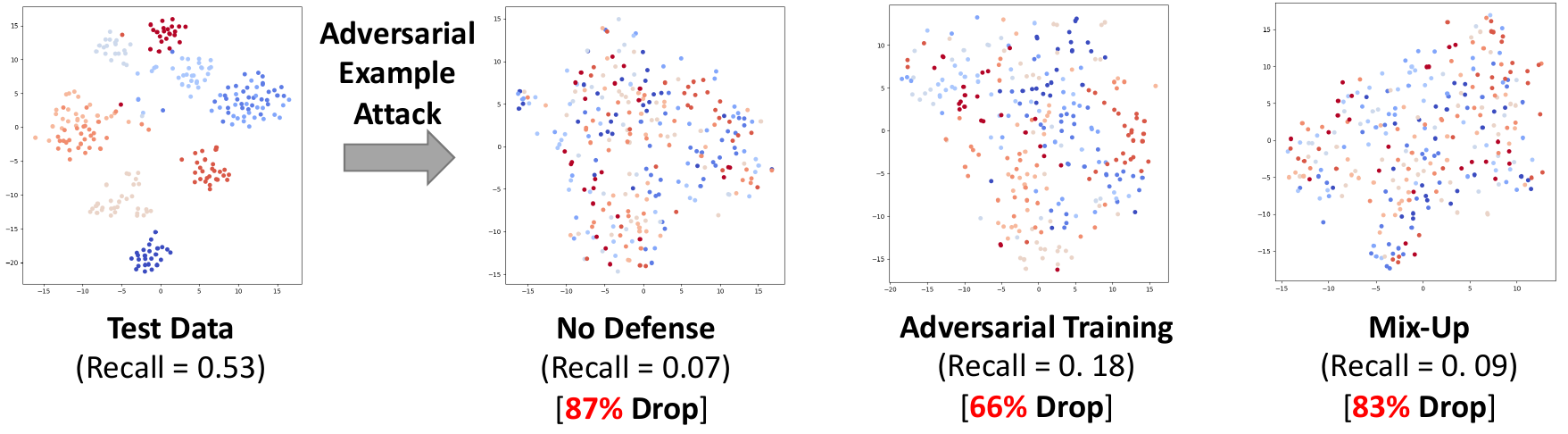}
\end{center}
\caption{The t-SNE \cite{Poliar731877} results of the adaptions of popular defenses (Adversarial Training~\cite{madryDeepLearningModels2019a} and Mix-Up~\cite{zhangMixupEmpiricalRisk2018}. The DML models are trained on CUB200 \cite{WelinderEtal2010} using the Proxy Anchor Loss \cite{kimProxyAnchorLoss2020}. Before launching the attack, the DML model can clearly classify samples from different classes. But after the attack, DML models trained under different settings all showed poor performance.}
\label{fig:tsne-cub200}
\end{figure}

The issue of adversarial examples is a thorn for deep neural networks. \citeauthor{szegedyIntriguingPropertiesNeural2014} \shortcite{szegedyIntriguingPropertiesNeural2014} have first shown the robustness vulnerability of deep neural networks. They found that adding a specially crafted imperceptible perturbation on a natural image could greatly affect the prediction of a deep classification model. These perturbed images are called adversarial examples. Since then, there has been an arms race between attackers and defenders on this issue. However, this arms race is mainly about deep classification models, leaving DML models largely unexplored.

There are few existing works focus on the robustness of DML models \cite{baiAdversarialMetricAttack2020, panumExploringAdversarialRobustness2021}, but those works failed to cover all the important applications of DML models. The use of trained DML models can be categorized into template matching and clustering. As to template matching, for a given input sample, the DML model extracts an embedding vector and then computes the distance (e.g, $l_2$-norm ) from that vector to each template embedding. The label of the template with the shortest distance is used as the prediction result \cite{shi2015constrained, xiao2017joint, wojke2018deep, huang2021full}.
As for clustering, there is no prior-known template embedding. Clustering is performed over the embeddings of a set of unlabeled samples  \cite{oh2017deep,wang2017deep,chen2018almn,duanDeepAdversarialMetric, chenEnergyConfusedAdversarial2019, chenHybridAttentionBasedDecoupled2019, duttaUnsupervisedDeepMetric2020,huang122020marginalized,hu2021learning}.
Existing works \cite{baiAdversarialMetricAttack2020, panumExploringAdversarialRobustness2021} all fall into the template matching category but neglect the clustering-based use. Defending against adversarial examples for the clustering-based scenarios is much harder than that of template matching scenarios. For template matching scenarios, in each distance computing, the attacker has the chance to manipulate one input term, the embedding vector, by adding perturbations on the corresponding sample, but does not have the chance to modify the other input term, the template embedding, which is pre-calculated and fixed during the inference. However, in each distance computing of the clustering scenario, the attacker has the chance to manipulate both the two input terms by adding perturbations on the two corresponding samples, and that means the attacker has a bigger space to exploit, making the defense much harder.

Also, existing defenses designed for the template matching scenarios are not suitable for the more challenging clustering scenarios. Existing defenses adapted adversarial training methods design for deep classification models to DML scenarios. Since those methods are designed without considering clustering scenarios, their adversarial training processes require labeled templates that are unavailable in the clustering scenario, making those methods unsuitable.

The design principles of defenses designed for deep classification models are unsuitable for the scenario considered by us either because DML models have unstable label vectors and their embedding spaces are high-dimensional (Details will be covered in the Problem Formulation Section). Figure \ref{fig:tsne-cub200} shows the results of directly using some adaptions as defenses (More results are in Table~\ref{tab:mobi-cub200-cars196} and Table~\ref{tab:bn-cub200-cars196}. Those results suggest that adaptions cannot achieve strong robustness for the clustering scenarios.

In this paper, we propose a new method, the Ensemble Adversarial Training (EAT), to enhance the adversarial robustness of DML. EAT fuses ensemble learning and adversarial training considering the training characteristics of DML. Instead of exploiting the uninterpretable high-dimensional embedding space \cite{pangImprovingAdversarialRobustness2019}, to promote the diversity of the ensemble, EAT leverages the training data arrangement, which is not only lightweight but also friendly to the training of DML models. EAT also employs a self-transferring mechanism to distribute the robustness statistics of the whole ensemble to each model in the ensemble to further enhance the model robustness.

We evaluate the proposed method on three mostly used DML datasets (CUB200 \cite{WelinderEtal2010}, CARS196 \cite{KrauseStarkDengFei-Fei_3DRR2013}, and In-Shop \cite{liuLQWTcvpr16DeepFashion}) with two popular models (MobileNet-V2 \cite{DBLP:journals/corr/abs-1801-04381} and BN-Inception \cite{DBLP:journals/corr/IoffeS15}), using the state-of-the-art metric learning design. And the results show that our method greatly outperforms the defenses which are adapted from methods designed for deep classifcation models.

The contributions of this paper are summarized below:
\begin{itemize}
    \item To the best of our knowledge, we are the first to point out the threat of adversarial examples in the clustering-based inference of DML models, and show that the adaptions of defenses originally designed for deep classification models cannot work well for this DML scenario.
    \item We propose the EAT method as a defense for enhancing the robustness of DML models used in clustering scenarios. The EAT method promotes the model diversity of the ensemble by exploiting the training data arrangement, reducing the chance that all the models in the ensemble are simultaneously fooled by one sample. Additionally, the EAT method also uses a self-transferring mechanism to make full use of the ensemble's robustness statistics.
    \item We evaluate the EAT method on three popular datasets (CUB200, CARS196, and In-Shop) with two models (MobileNetV2 and BN-Inception). Our results show that our EAT method greatly outperforms other methods adapted from defenses originally designed for deep classification tasks in terms of model robustness while introducing a negligible performance drop on the original DML task.
\end{itemize}

The rest of this paper is organized as follows. First, we introduce related works and some background knowledge in the Related Works Section. Then, we formalize our problem and discuss adapted forms of famous defenses originally designed for deep classification tasks (Problem Formulation Section). Next, we describe our proposed EAT method in the Proposed Approach Section. Finally, we show our experimental settings and evaluation results in the Experiments Section.

\section{Related Works}

\subsection{Deep Metric Learning}
DML aims to learn a distance representation for each input through deep neural networks \cite{parkhi2015deep, schroff2015facenet}. Similar inputs should produce close representations under the same metric learning model. The key of DML is the loss function, which is evaluated using a tuple consisting of samples from the same class and samples from different classes (The formal definitions will be covered in the Problem Formulation Section).
There are many famous loss designs for metric learning, including the contrastive loss \cite{bromley1994signature, chopra2005learning, hadsell2006dimensionality} and the triplet loss \cite{schroff2015facenet}. 

The contrastive loss and the triplet loss are pair-based losses, and those kinds of losses need to sample data and build tuples in the model training. The sampling operation in pair-based loss introduces high complexity. To reduce the complexity of the pair-based loss, the proxy-based methods \cite{aziere2019ensemble} are proposed. The proxy-based methods always maintain proxies to represent each class, and the corresponding loss is computed using those proxies and real training samples. Also, \citeauthor{kimProxyAnchorLoss2020} \shortcite{kimProxyAnchorLoss2020} propose a new proxy-based loss called PAL by making a combination of the SoftTriple Loss \cite{qianSoftTripleLossDeep2020} and the Proxy-NCA Loss \cite{movshovitz2017no}. In this paper, we select the PAL which is the state-of-the-art as our deep metric loss.

\subsection{Adversarial Examples for Classification Models}

\subsubsection{Attacks}

Initially, \citeauthor{szegedyIntriguingPropertiesNeural2014} \shortcite{szegedyIntriguingPropertiesNeural2014} found that deep neural networks can be fooled by some specific perturbations on the input. 
 \citeauthor{goodfellowExplainingHarnessingAdversarial2015} \shortcite{goodfellowExplainingHarnessingAdversarial2015} then proposed Fast Gradient Sign Method (FGSM). FGSM exploits the sign of gradients resulted from the backpropagation to construct perturbations. 
 \citeauthor{papernotLimitationsDeepLearning2015} \shortcite{papernotLimitationsDeepLearning2015} designed the Jacobian-based Saliency Map Attack (JSMA) through a saliency map based on the gradient. 
\citeauthor{carliniEvaluatingRobustnessNeural2017} \shortcite{carliniEvaluatingRobustnessNeural2017} developed the C\&W attack. That attack alleviates the linear property in the optimization which is used for searching the perturbation.  \citeauthor{madryDeepLearningModels2019a} \shortcite{madryDeepLearningModels2019a} proposed Projected Gradient Method (PGD) to further improve the attack performance by the project operation.


In this work, we follow common practice~\cite{pangImprovingAdversarialRobustness2019,pangMixupInferenceBetter2020} and select the PGD attack to evaluate the robustness of deep learning models.

\subsubsection{Defenses}
There are many defenses proposed to improve the adversarial robustness of the classification learning model against adversarial examples. Those defenses can be categorized as training stage defenses~\cite{goodfellowExplainingHarnessingAdversarial2015, zhangMixupEmpiricalRisk2018, vermaManifoldMixupBetter2019, madryDeepLearningModels2019a} and inference stage defenses~\cite{xie2017mitigating, pangMixupInferenceBetter2020}.


Training stage defenses. \citeauthor{madryDeepLearningModels2019a} \shortcite{madryDeepLearningModels2019a} propose a defense, the Adversarial Training (AT), which dynamically generates adversarial examples during the model training and uses those generated samples as training samples. AT-based methods have shown superior performance over other kinds of defenses on defending against adversarial examples.
\citeauthor{lambInterpolatedAdversarialTraining2019} \shortcite{lambInterpolatedAdversarialTraining2019} propose a defense called Interpolated Adversarial Training (IAT) which combines the Mix-Up methods with AT methods.
\citeauthor{pangMixupInferenceBetter2020} \shortcite{pangMixupInferenceBetter2020} propose a defense based on the Mix-Up training paradigm. In the model training, they directly perform linear combinations to generate new training samples instead of launching adversarial attacks as AT does.

Inference stage defenses. \citeauthor{guo2017countering} \shortcite{guo2017countering}, \citeauthor{xie2017mitigating} \shortcite{xie2017mitigating} and \citeauthor{raff2019barrage} \shortcite{raff2019barrage} propose methods based on the similar design principle that performs a non-linear transformation on the input of deep learning models.

\subsection{Adversarial Examples for DML}




There are few works that are designed for defending DML models against adversarial examples~\cite{baiAdversarialMetricAttack2020, panumExploringAdversarialRobustness2021}.
Those works only consider the template matching use of DML models (Recall the template matching based DML inference and clustering-based inference described in the Introduction Section). The defense methods of those works follow an adversarial training manner and require labeled templates for generating adversarial examples in the model training. However, there is no template in the clustering-based inference of DML models, and that suggests those defenses cannot be used for the clustering-based inference.


\section{Problem Formulation}

\subsection{Notations}
We denote the DML model as $\mathcal{F}(x;\theta)$, which takes an $m$-dimensional sample $x\in \mathbb{R}^m$ as the input and $\theta$ as model parameters. The output of $\mathcal{F}(\cdot)$ is an embedding vector with $d$ dimensions ($\mathcal{F}(x;\theta) = \hat{y} \in \mathbb{R}^d$). We denote the training dataset and the test dataset as $D_{train} = \{(x_1,y_1), (x_2,y_2), ..., (x_{N},y_{N})\}$ and $D_{test} = \{(x_1,y_1), (x_2,y_2), ..., (x_{M},y_{M})\}$, respectively. $N$ is the number of training samples and $M$ is the number of test samples. We represent the adversarial example of a specific input $x$ as $x_{adv} = x + \delta$,  where $\delta$ is the perturbation computed by the attacker and usually has small amplitude. Following the same choice in \cite{pangMixupInferenceBetter2020, zhangMixupEmpiricalRisk2018, madryDeepLearningModels2019a, vermaManifoldMixupBetter2019}, in this paper, we use $l_p$-norm to bound $\delta$ and consider the $\delta$ s.t. $||\delta||_p \leq \varepsilon$ as valid perturbations.

\subsection{Problem Definition}
%

Before introducing the formal definition of the problem, we first review the definition for defending against adversarial examples for traditional classification tasks, which can be written as:
\begin{equation}
	\min_{\theta} \mathbb{E}_{(x,y)\sim D}\Big[\max_{x_{adv} \in S(x, \varepsilon)}\mathcal{L}(\mathcal{C}(x_{adv};\theta), y)\Big]
	\label{eq:cls-goal}
\end{equation}

$S(x, \varepsilon)$ is the set of possible adversarial examples generated using $x$ as the starting point for attacking classification model $\mathcal{C}(\cdot)$. And $\mathcal{L}(\cdot)$ is the loss function (e.g. cross-entropy loss) that evaluates the distance between the one-hot label ($y$) of $x$ and the output of classification model $\mathcal{C}(\cdot)$ when its input is $x_{adv}$. 
Given an input $x$, Equation~\ref{eq:cls-goal}  encourages $\mathcal{C}(\cdot)$ to output the correct label for all $x_{adv} \in S(x, \varepsilon)$, and thus making sure that the attacker can hardly find the small perturbation $\delta$ that can fool the classifier.

However, Equation~\ref{eq:cls-goal}  cannot be used for DML because there is no explicit label ($y$) for the data samples in DML tasks. DML is designed to model the distance over data samples. Specifically, DML extracts embedding vectors from input data, and then treats data samples whose embedding vectors have small distances as the same class, while considers data samples whose embedding vectors have large distances as different classes. The label vector of a data sample is usually the centroid of the class it belongs to in the embedding space and is unclear untill the training is finished.
  

The loss function of DML always depends on several samples instead of one sample, and its general form can be well represented by the triplet loss~\cite{schroff2015facenet}:
\begin{equation}
	\begin{aligned}
		&\mathcal{L}_{metric}(\mathcal{F}(x), \mathcal{F}(x^+), \mathcal{F}(x^-)) = \\ &||\mathcal{F}(x)-\mathcal{F}(x^+)||
		+ \max(0, m - ||\mathcal{F}(x)-\mathcal{F}(x^-)||)
	\end{aligned}
\end{equation}
where $x$, $x^+$ are data samples of the same class while $x^-$ is from another class, and  $m$ is a constant number. Similarly to Equation~\ref{eq:cls-goal}, we can write the problem of defending against adversarial examples for DML as:
\begin{equation}
\begin{aligned}
    & \min_{\theta}  \mathbb{E}_{(x,x^+,x^-) \sim D}  \Big[ \max_{
        x_{adv},x^+_{adv},x^-_{adv}\in S(x, x^+, x^-, \varepsilon) 
    } \\& \mathcal{L}_{metric}(\mathcal{F}(x_{adv};\theta),\mathcal{F}(x^+_{adv};\theta),\mathcal{F}(x^-_{adv};\theta)) \Big]
\end{aligned}
\label{eq:metric-goal}
\end{equation}
where $S(x, x^+, x^-, \varepsilon)$ is the abbreviation of $\{S(x,\varepsilon), S(x^+, \varepsilon), S(x^-, \varepsilon) \}$.

\subsection{Previous Defenses} \label{sec:previous_defense}
In this section, we will introduce the definition of four kinds of representative defenses designed for classification models and analyse whether or not their design principles fit for DML. We give adapted forms of these defenses for DML. We also empirically show that those defenses are unsuitable for DML tasks in the Experiments Section.

\textbf{Adversarial Training (AT) \cite{madryDeepLearningModels2019a}. }  AT improves the adversarial robustness by introducing adversarial examples into the training phase. At each training iteration, AT generates adversarial examples based on the current model and includes those examples into the training dataset. AT minimizes the empirical risk as:
\begin{equation} \label{eq:adversarial_example_for_classification}
    \min_{\theta} \frac{1}{|D|}\sum_{(x,y)\in D}\mathcal{L}(\mathcal{C}(x_{adv}),y) 
\end{equation}
where $x_{adv}$ is generated by launching adversarial attacks. Taking the most powerful and representative attack PGD \cite{madryDeepLearningModels2019a} as an example, generating adversarial examples can be described as:

\begin{equation} \label{eq:pgd_for_classification}
    x_{adv}^{(t+1)} = \text{Proj}_{x +\mathcal{S}}\left(x_{adv}^{(t)} + \alpha \text{sgn}(\nabla_x \mathcal{L}(\mathcal{C}(x_{adv}^{(t)}),y))\right)
\end{equation}
where $\text{Proj}_{x+\mathcal{S}}(z)$ is the projection function which can project $z$ into the space $x+\mathcal{S}$ and $\mathcal{S}$ is the possible adversarial example space of $x$. 

For DML, Equation~\ref{eq:adversarial_example_for_classification} and \ref{eq:pgd_for_classification} cannot be directly used because there is no fixed label vector  $y$ during the model training. To tackle this problem, we use $\mathcal{F}(\cdot)$ as the replacement of $y$ and use $\mathcal{L}_{metric}(\cdot)$ and $\mathcal{F}(\cdot)$ in place of $\mathcal{L}(\cdot)$ and $\mathcal{C}(\cdot)$, respectively. It is worth noting that the dynamic ``label'' $\mathcal{F}(x)$ could make the training hard to converge.



\textbf{Mix-Up Training \cite{zhangMixupEmpiricalRisk2018}.}
Mix-Up Training methods have shown great performance in improving the adversarial robustness of deep neural networks. These kinds of methods have lower complexity compared to the AT-based methods. The target of the Mix-Up Training methods can be written as:
\begin{equation} \label{eq:adversarial_training_mix_up}
    \min_{\theta} \frac{1}{m} \sum_{i=1}^{m}{\mathcal{L}(\mathcal{C}(\tilde{x}_i), \tilde{y}_i)}
\end{equation}
where $\tilde{x}_i = \lambda x_{i0} + (1-\lambda)x_{i1}$, $\tilde{y}_i = \lambda y_{i0} + (1-\lambda)y_{i1}$, $\lambda \sim Beta(\alpha, \alpha)$ and $m$ is the number of training samples. Mix-Up Training methods exploit the linear combination of training sample instead of generating adversarial examples in the model training, which obtains lower computational complexity. 

To make Mix-up training available for DML, we propose to replace the $\tilde{y}_i$ in Equation~\ref{eq:adversarial_training_mix_up} as $\lambda\mathcal{F}(x_{i0}) + (1-\lambda)\mathcal{F}(x_{i1})$, and replace  the  $\mathcal{L}(\cdot)$ and $\mathcal{C}(\cdot)$ with $\mathcal{L}_{metric}(\cdot)$ and $\mathcal{F}(\cdot)$, respectively.
Since the new label vector $\tilde{y}_i = \lambda\mathcal{F}(x_{i0}) + (1-\lambda)\mathcal{F}(x_{i1})$ changes as the update of $\mathcal{F}(\cdot)$ and is highly dynamic, the training could be greatly effected and be hard to converge.

\textbf{Interpolated Adversarial Training (IAT) \cite{lambInterpolatedAdversarialTraining2019}}. IAT is a fusion of the Mix-Up Training and  AT. The target of IAT can be written as:
\begin{equation}
    \min_{\theta} \frac{1}{m} \sum_{i=1}^{m}{\mathcal{L}(\mathcal{C}(\tilde{x}_i), \tilde{y}_i)} + \frac{1}{m} \sum_{i=1}^{m}{\mathcal{L}(\mathcal{C}(g(\tilde{x}_i)), \tilde{y}_i)} 
\end{equation}
where $g(\tilde{x}_i) = \lambda Adv(x_{i0}) + (1-\lambda) Adv(x_{i1})$, $Adv(\cdot)$ is an arbitrary adversarial example attack and $\lambda \sim Beta(\alpha, \alpha)$. We design to adapt IAT to DML by using the same method used in adapting Mix-Up training. Due to the similar design, the IAT method may also have the convergence issue as Mix-up training does under DML scenarios.



\textbf{TRADES \cite{zhangTheoreticallyPrincipledTradeoff2019}}. TRADES is a defensive method that exploits the trade-off between robustness and accuracy. The target of TRADES can be written as:
\begin{equation}
    \min_{\theta} \frac{1}{m} \sum_{i=0}^{m} \big( \mathcal{L}(C(x_i),y_i) + \lambda \cdot\mathcal{L}(C(Adv(x_i)),C(x_i)) \big)
\end{equation}
where $Adv(\cdot)$ is an arbitrary adversarial example attack and $\lambda > 0$ is a hyper-parameter weighting the importance of the two terms. TRADES has demonstrated great performance on classification tasks, but it still relies on the fixed one-hot target vector, which may lead to convergence issues when it is adapted to DML scenarios. 
We can naturally change the first loss term to the metric loss and replace the $C(\cdot)$ with $\mathcal{F}(\cdot)$ to enable this method for DML. 

\textbf{Ensemble Training}.
\citeauthor{pangImprovingAdversarialRobustness2019} \shortcite{pangImprovingAdversarialRobustness2019} propose a defense by enlarging the diversity of the ensemble to improve the adversarial robustness. They define the ensemble diversity as $\mathbb{ED} = \textit{det}(M^TM)$ where $M = (\mathcal{F}^{(1)}(x), \mathcal{F}^{(2)}(x), ..., \mathcal{F}^{(K)}(x))\in \mathbb{R}^{(L-1)\times K}$ and $\mathcal{F}^{(i)}$ is the order-preserving prediction of the $i$-th classifier on the input $x$. The training goal of their defense is to maximize the diversity term $\mathbb{ED}$ to 1. 
However, this training goal could be difficult to achieve when the output of $\mathcal{F}^{(i)}$ is high-dimensional. Unfortunately, DML models usually have a large embedding space to obtain better performance, making that ensemble diversity design unsuitable.


\textbf{Previous defenses cannot work well for DML}.
We conduct experiments (in the Experiments Section) to verify the effectiveness of those adapted defenses. The results show that defense principles designed for deep classification models cannot work well in DML scenarios (Details are in Table~\ref{tab:mobi-cub200-cars196} and Table~\ref{tab:bn-cub200-cars196}), which suggests that we need new defensive designs dedicated for DML tasks. 


\section{Proposed Approach} \label{sec:approach}
In this section, we will propose a new method, the Ensemble Adversarial Training (EAT), for defending against adversarial examples to improve the robustness of DML models.

As discussed in the Problem Formulation Section, defensive methods designed for classification models cannot be directly used for DML and their adapted forms also cannot work well because of unstable label vectors and the high-dimensional issue.

To avoid those issues, we propose a new method based on ensemble training and adversarial training. Different from the method proposed by \citeauthor{pangImprovingAdversarialRobustness2019} \shortcite{pangImprovingAdversarialRobustness2019}, we leverage the distribution difference of training data to promote the diversity of the ensemble instead of using the $\textit{det}(\cdot)$ operation that takes model output vectors as inputs.
Additionally, we construct a self-transferring mechanism that aims to further enhance the robustness of each model in the ensemble by considering the robustness statistics of the whole ensemble.

Algorithm \ref{alg:eat} shows our overall design. We randomly split the training set into
$N$ parts and the combinations that consists of $N-1$ parts are used as candidate training sets. Then, we assign a unique candidate training set to each model in the ensemble as its training set. This design is to promote the diversity of the ensemble.
Next, we start the model training that includes a normal metric learning part and an adversarial training part.

As to the normal metric learning part, each model evaluates its loss on its training set normally (Line 6 in Algorithm~\ref{alg:eat}). As for the adversarial training part, we use a self-transferring mechanism. Specifically, for the training of each model in the ensemble, we generate adversarial examples on the whole ensemble instead of that single model (Line 7) and those generated examples are later used in the training of that model (Line 8). This design shares the robustness statistics of the whole ensemble with each model and could further enhance the model robustness.

\renewcommand{\algorithmicrequire}{\textbf{Input:}}  
\renewcommand{\algorithmicensure}{\textbf{Output:}} 
\begin{algorithm}
    \caption{Ensemble Adversarial Training}
    \begin{algorithmic}[1]
    \REQUIRE $N$ individual models: $\mathcal{F}^i_{\theta^i}$ where $1\leq i \leq N$, Training Data: $D_{train} = \{(x_i,y_i)\}_{i=1}^{|D_{train}|}$,
             Epochs: $T$, perturbation budget: $\varepsilon$, learning rate: $\gamma$, mixed ratio: $\beta$
    \ENSURE Updated parameters: $\theta^{i}_{T}$
    \STATE Initialize the parameters $\theta^i = \theta^i_0$
    \STATE Splitting the training dataset into $N$ parts $D_{train} = \{D_{train}^1, D_{train}^2, ..., D_{train}^N\}$ with the same size 
    \FOR{$ep=1$ to $T$}
    \FOR{$i=1$ to $N$}
    \STATE $D_{ml} \leftarrow D_{train} \backslash \{D_{train}^i\}$
    \STATE $\mathcal{L}_{ml} \leftarrow \mathcal{L}_{metric}(D_{ml}, \mathcal{F}^i_{\theta^i_{ep-1}})$
    \STATE $D_{ml}^{adv} \leftarrow gen(D_{ml}, (\mathcal{F}^1_{\theta^i_{ep-1}}, \mathcal{F}^2_{\theta^i_{ep-1}}, ..., \mathcal{F}^N_{\theta^i_{ep-1}}), \varepsilon)$
    \STATE $\mathcal{L}_{adv} \leftarrow \mathcal{L}_{metric}(D_{ml}^{adv}, \mathcal{F}^i_{\theta^i_{ep-1}})$
    \STATE $\mathcal{L}_{EAT} \leftarrow \beta*\mathcal{L}_{ml} + (1-\beta)*\mathcal{L}_{adv}$
    \STATE $\theta_{ep}^i \leftarrow \theta_{ep-1}^i - \gamma*\nabla_{\theta_{ep-1}^i}\mathcal{L}_{EAT}$  
    \ENDFOR
    \ENDFOR
    \end{algorithmic}
    \label{alg:eat}
\end{algorithm}



\textbf{More Details about the generation of adversarial examples for adversarial training.} The generation function $gen(\cdot,\cdot,\cdot)$ in Line 7 of Algorithm \ref{alg:eat} can be formulated as:
\begin{equation}
\begin{aligned}
gen(D,\mathcal{F},\varepsilon) &= \Big\{p(x, \mathcal{F}, \varepsilon)| x \in D \Big\} \\
\mathcal{F} &= (\mathcal{F}^1,\mathcal{F}^2,...,\mathcal{F}^N)
\end{aligned}
\end{equation}
$p(\cdot,\cdot,\cdot)$ is a variant of the PGD attack and its iterative form can be written as:
\begin{equation}
\begin{aligned}
    x_{adv}^{(t+1)} = clip\Big(\text{Proj}_{x+\mathcal{S}} \big(x_{adv}^{(t)} + \alpha\text{sgn}(\mathbb{SG})\big), \varepsilon\Big)\\
    \mathbb{SG} = \sum_{i=1}^{N}\nabla_x\mathcal{L}\big(\mathcal{F}^{(i)}(x_{adv}^{(t)}), \mathcal{F}^{(i)}(x)\big)\\
\end{aligned}
\end{equation}
where $x_{adv}^{(0)} = x$, $0<t<T$, $T$ is the maximum number of attack iterations, and $\mathbb{SG}$ denotes the sum of all the gradients. Only the samples generated in the last iteration are kept ($p(x, \mathcal{F}, \varepsilon) = x_{adv}^{(T)}$).


\textbf{Model inference.} 
After the EAT training, we will obtain a bunch of models. For the model inference, we need a method that not only leverages the knowledge of the whole ensemble but also is friendly to model robustness. Here, we denote the output of the ensemble as $(e_1,e_2,...,e_N)$ for a given input $x$. A possible choice for model inference is to average all the output vector of the ensemble as the new output, i.e. $e = \frac{1}{N}\sum_{i=1}^{N}e_i$.
This method is simple, though, the drawback is that the averaging operation is differentiable which might give the attacker an advantage for launching attacks. So we use an undifferentiable voting mechanism for the inference. Specifically, for a given input $x$, we count the predicted label of each model in the ensemble and select the label that appears most frequently as the prediction result.

\section{Experiments}

In this work, we consider the adversarial example attack in the DML tasks under the white-box setting (which means the attacker knows the structure and the parameters of the deep learning model). In this section, we first evaluate our EAT defense on three widely used datasets (CUB200, CARS196, and In-Shop) with two popular models (MobileNetV2 and BN-Inception). We then analyze the performance under various settings.

\subsection{Datasets}
We evaluate our methods on CUB200 \cite{WelinderEtal2010}, CARS196 \cite{KrauseStarkDengFei-Fei_3DRR2013} and In-Shop \cite{liuLQWTcvpr16DeepFashion}. CUB200 is an image dataset with photos of 200 bird species. For CUB200, we select its first 100 classes which contain 2944 images as our training set and other 3089 images of the other classes as our test set. CARS196 is an image dataset with 196 classes of cars. For the CARS196, we choose the 8054 images of its first 98 classes as our training set and the other 8131 images are used for testing. For the In-Shop, we use its first 3997 classes which contain 25882 images for training and the other 28760 images are used for testing. The test set of the In-Shop consists of three parts: the training part, the query part and the gallery part. The training part contains 25882 images. The query part contains 14218 images of 3986 classes and the gallery part has 12612 images of 3985 classes. In this work, we only select the query part as the test data to analyze the robustness like the test part of CUB200 and CARS196.

\subsection{Models and Metric Learning Loss}
In this work, we select the MobileNetV2 \cite{DBLP:journals/corr/abs-1801-04381} and the BN-Inception \cite{DBLP:journals/corr/IoffeS15} as the deep learning model to train and test. In every experiment, we all use the corresponding pre-trained model on the ImageNet \cite{imagenet_cvpr09} as the initial model before the training phase. 
For the metric learning loss, since the PAL loss~\cite{kimProxyAnchorLoss2020} achieves the SOTA performance, we choose it as the metric learning loss, i.e. $\mathcal{L}_{metric} = \mathcal{L}_{pal}$.

\subsection{Evaluation Metrics}
In the evaluation phase, we exploit three metrics (Recall, F1-Score, and NMI) to evaluate the performance of the deep learning model in DML tasks.

The form of the recall value can be written as:
\begin{equation}
    Recall@k = \frac{TP}{TP+FN}
\end{equation}
where $TP$ is the true-positive counts and the $FN$ is the false-negative counts in top-$k$ predictions. 

The form of the F1-Score is:
\begin{equation}
    \text{F1-Score} = \frac{2TP}{2TP+FN+FP}
\end{equation}
where $FP$ is the false-positive counts. The F1-Score is the combination of recall and precision.

The NMI value is the Normalized Mutual Information. The form of the NMI can be written as:
\begin{equation}
    NMI(\Omega, C) = \frac{2\mathbb{I}(\Omega, C)}{H(\Omega)+H(C)}
\end{equation}
where $\Omega$ is the class label, $C$ is the cluster label, $\mathbb{I}(\cdot, \cdot)$ is the mutual information function and $H(\cdot)$ is the entropy function.

\subsection{Implementation Details}

During the training step, we exploit the AdamW \cite{DBLP:journals/corr/abs-1711-05101} optimizer which has the same update process of the Adam \cite{kingma2014adam} yet decays the weights separately to perform the backpropagation.
Our models are trained for 200 epochs with the initial learning rate $10^{-4}$ on three datasets. We apply the learning rate scheduler to adjust the learning rate during the training
phase. During the training phase, we set the PGD attack with $\varepsilon = 16/255$ and the iteration number is set as 10. In the test phase, we set $\varepsilon = 8/255$ and the iteration number is also set as 10. As for the implementation, our code is based on the open-source code of \citeauthor{roth2020revisiting} \shortcite{roth2020revisiting} and \citeauthor{pangMixupInferenceBetter2020} \shortcite{pangMixupInferenceBetter2020}.





\subsection{Result Analysis}

\textbf{The adversarial robustness comparison of all defenses.} As is shown in Table \ref{tab:mobi-cub200-cars196} to Table \ref{tab:mob-in-shop}, five kinds of defenses all obtain great performance under the clean mode (i.e. without any attacks). These defenses all cause a little drop of all metrics after the training phase. The AT method presents minimum performance drop under the clean mode and the Mix-Up method shows maximum drop without any attacks. As for the adversarial robustness of these defenses, the Mix-Up method and the IAT method fail under the PGD attack with 10 iterations with a huge drop of all the metrics. Our proposed EAT defense presents the SOTA performance under the PGD attack. Especially on the recall metric, our EAT defense significantly outperforms other defenses.

\begin{table*}[]
\begin{center}
\begin{adjustbox}{width=((\columnwidth)*2),center}
\begin{tabular}{cc|cccccc|cccccc}
\hline
Method                                 &                       & NMI            & F1-Score      & Recall@1       & Recall@2       & Recall@4       & Recall@8       & NMI            & F1-Score      & Recall@1       & Recall@2       & Recall@4       & Recall@8       \\ \hline
                                       & \multicolumn{1}{l|}{} & \multicolumn{6}{c|}{CUB200}                                                                        & \multicolumn{6}{c}{CARS196}                                                                       \\ \hline
\multirow{2}{*}{\textbf{PAL}}          & Clean                 & 59.97          & 22.73         & 53.27          & 64.15          & 74.47          & 84.03          & 51.02          & 16.16         & 51.40          & 63.52          & 74.59          & 84.73          \\ \cline{2-14} 
                                       & PGD-10                & 30.95          & 1.73          & 7.16           & 10.84          & 16.17          & 23.89          & 27.10          & 1.97          & 5.52           & 8.89           & 14.22          & 23.09          \\ \hline
\multirow{2}{*}{\textbf{PAL + AT}}     & Clean                 & 59.52          & 21.15         & 53.24          & 63.52          & 73.71          & 82.68          & 48.29          & 14.18         & 39.97          & 52.60          & 64.87          & 76.28          \\ \cline{2-14} 
                                       & PGD-10                & 35.54          & 3.58          & 17.98          & 23.89          & 32.53          & 43.12          & 31.72          & 4.18          & 14.15          & 20.28          & 28.68          & 38.80          \\ \hline
\multirow{2}{*}{\textbf{PAL + Mix-Up}} & Clean                 & 56.70          & 18.58         & 51.66          & 62.34          & 72.92          & 82.29          & 47.89          & 12.89         & 45.00          & 58.15          & 69.58          & 80.33          \\ \cline{2-14} 
                                       & PGD-10                & 31.32          & 2.01          & 8.77           & 12.88          & 18.86          & 26.85          & 28.11          & 2.20          & 7.26           & 11.85          & 18.47          & 28.83          \\ \hline
\multirow{2}{*}{\textbf{PAL + IAT}}    & Clean                 & 57.67          & 19.68         & 51.36          & 62.21          & 72.66          & 82.16          & 52.90          & 16.39         & 56.02          & 67.95          & 77.55          & 85.66          \\ \cline{2-14} 
                                       & PGD-10                & 32.27          & 2.39          & 12.32          & 17.78          & 24.55          & 33.49          & 31.35          & 3.62          & 14.41          & 21.60          & 30.49          & 40.80          \\ \hline
\multirow{2}{*}{\textbf{PAL + TRADES}} & Clean                 & 61.11          & 23.31         & 55.87          & 66.71          & 77.23          & 85.67          & 56.50          & 20.33         & 58.59          & 70.49          & 80.80          & 88.74          \\ \cline{2-14} 
                                       & PGD-10                & 31.54          & 2.33          & 10.81          & 15.22          & 21.03          & 29.21          & 26.17          & 1.93          & 6.57           & 10.48          & 16.05          & 24.26          \\ \hline
\multirow{2}{*}{\textbf{PAL + EAT}}    & Clean                 & 56.91          & 19.28         & 52.78          & 62.34          & 71.31          & 80.48          & 51.23          & 14.95         & 54.68          & 65.18          & 74.69          & 83.09          \\ \cline{2-14} 
                                       & PGD-10                & \textbf{38.02} & \textbf{4.84} & \textbf{24.02} & \textbf{30.43} & \textbf{37.56} & \textbf{47.49} & \textbf{36.08} & \textbf{5.77} & \textbf{22.80} & \textbf{31.52} & \textbf{42.46} & \textbf{54.21} \\ \hline
\end{tabular}
\end{adjustbox}
\end{center}
\caption{The comparison of five defenses on CUB200 and CARS196 with MobileNetV2 under the white-box adversarial example attack.}
\label{tab:mobi-cub200-cars196}
\end{table*}

\begin{table}[]
\begin{center}
\begin{adjustbox}{width=((\columnwidth)),center}
\begin{tabular}{cc|ccccc}
\hline
Method                                 &        & NMI            & F1-Score      & Recall@1       & Recall@2       & Recall@4       \\ \hline
\multirow{2}{*}{\textbf{PAL}}          & Clean  & 58.17          & 20.15         & 55.01          & 66.41          & 75.68          \\ \cline{2-7} 
                                       & PGD-10 & 31.88          & 2.35          & 11.53          & 16.60          & 23.83          \\ \hline
\multirow{2}{*}{\textbf{PAL + AT}}     & Clean  & 57.20          & 20.23         & 53.73          & 64.08          & 74.17          \\ \cline{2-7} 
                                       & PGD-10 & 36.17          & 4.25          & 21.75          & 28.03          & 36.61          \\ \hline
\multirow{2}{*}{\textbf{PAL + Mix-Up}} & Clean  & 58.67          & 20.68         & 55.14          & 66.09          & 75.48          \\ \cline{2-7} 
                                       & PGD-10 & 31.14          & 2.10          & 11.80          & 16.63          & 23.60          \\ \hline
\multirow{2}{*}{\textbf{PAL + IAT}}    & Clean  & 57.19          & 19.64         & 53.57          & 64.15          & 74.30          \\ \cline{2-7} 
                                       & PGD-10 & 36.42          & 4.49          & 21.10          & 28.16          & 35.75          \\ \hline
\multirow{2}{*}{\textbf{PAL + TRADES}} & Clean  & 58.68          & 21.07         & 55.37          & 65.63          & 76.31          \\ \cline{2-7} 
                                       & PGD-10 & 33.97          & 2.97          & 14.53          & 20.54          & 27.80          \\ \hline
\multirow{2}{*}{\textbf{PAL + EAT}}    & Clean  & 56.65          & 19.67         & 55.93          & 65.76          & 74.14          \\ \cline{2-7} 
                                       & PGD-10 & \textbf{38.42} & \textbf{4.89} & \textbf{25.50} & \textbf{32.67} & \textbf{40.32} \\ \hline
\end{tabular}
\end{adjustbox}
\end{center}
\caption{The comparison of five defenses on CUB200 with BN-Inception under the white-box adversarial example attack.}
\label{tab:bn-cub200-cars196}
\end{table}

\begin{figure}[]
\begin{center}
\includegraphics[width=\linewidth]{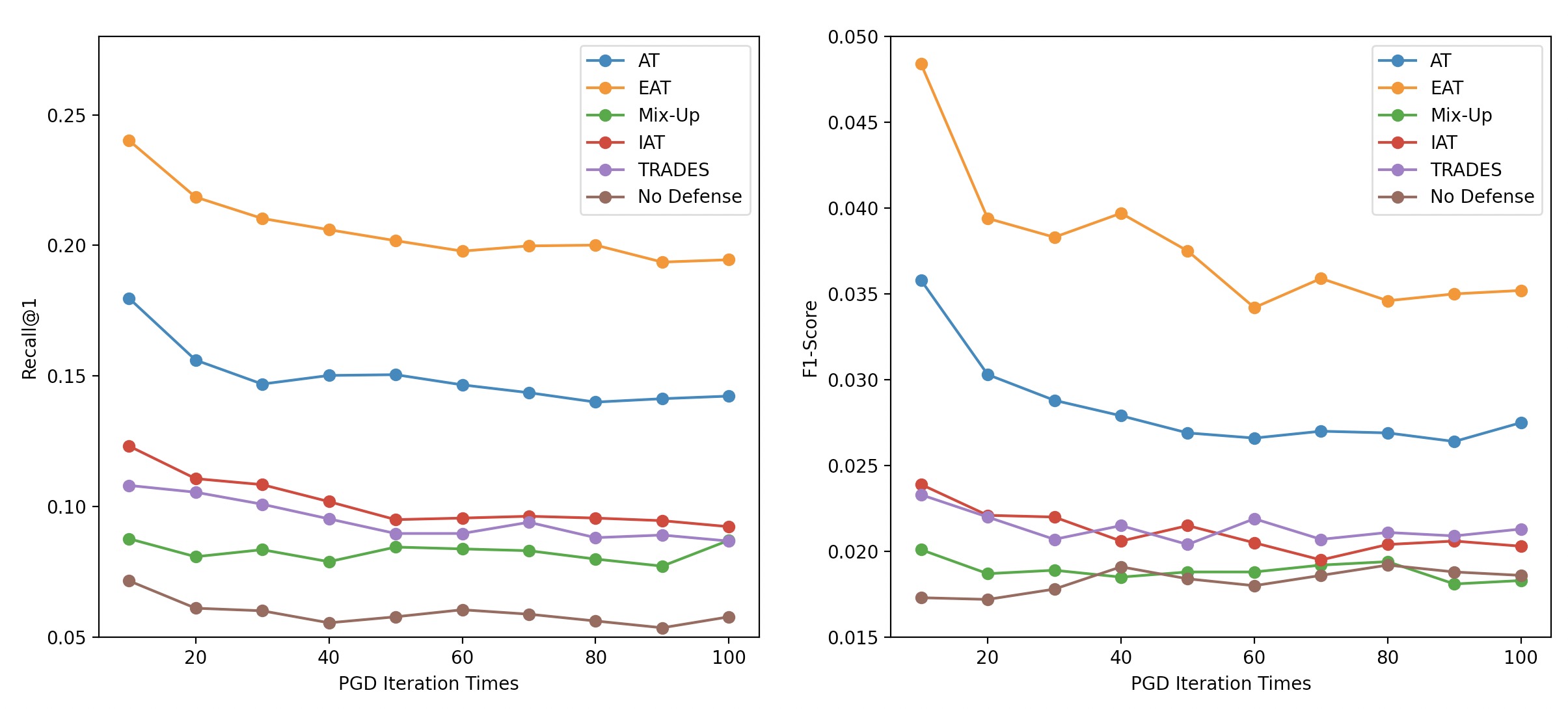}
\end{center}
  \caption{Recall@1 and F1-Score for different iteration numbers of the PGD attack with the different defenses on CUB200 (Training with MobileNetV2)}
\label{fig:iter-recall@1-f1-mobi-cub200}
\end{figure}

\textbf{The adversarial robustness under different iteration numbers of the attack.} As is shown in Figure \ref{fig:iter-recall@1-f1-mobi-cub200}, our proposed EAT defense presents the SOTA adversarial robustness than other defenses under the PGD attack with different iteration numbers.  
Under the PGD attack, the Mix-Up method is close to the No Defense. That means the effect of the Mix-Up method is small and it is not suitable for applying that kind of method in the DML tasks. The F1-Score and the Recall@1 decrease when we increasing the iteration number of the PGD attack but there are some fluctuations due to the randomness. 



\begin{figure}[]
\begin{center}
\includegraphics[width=\linewidth]{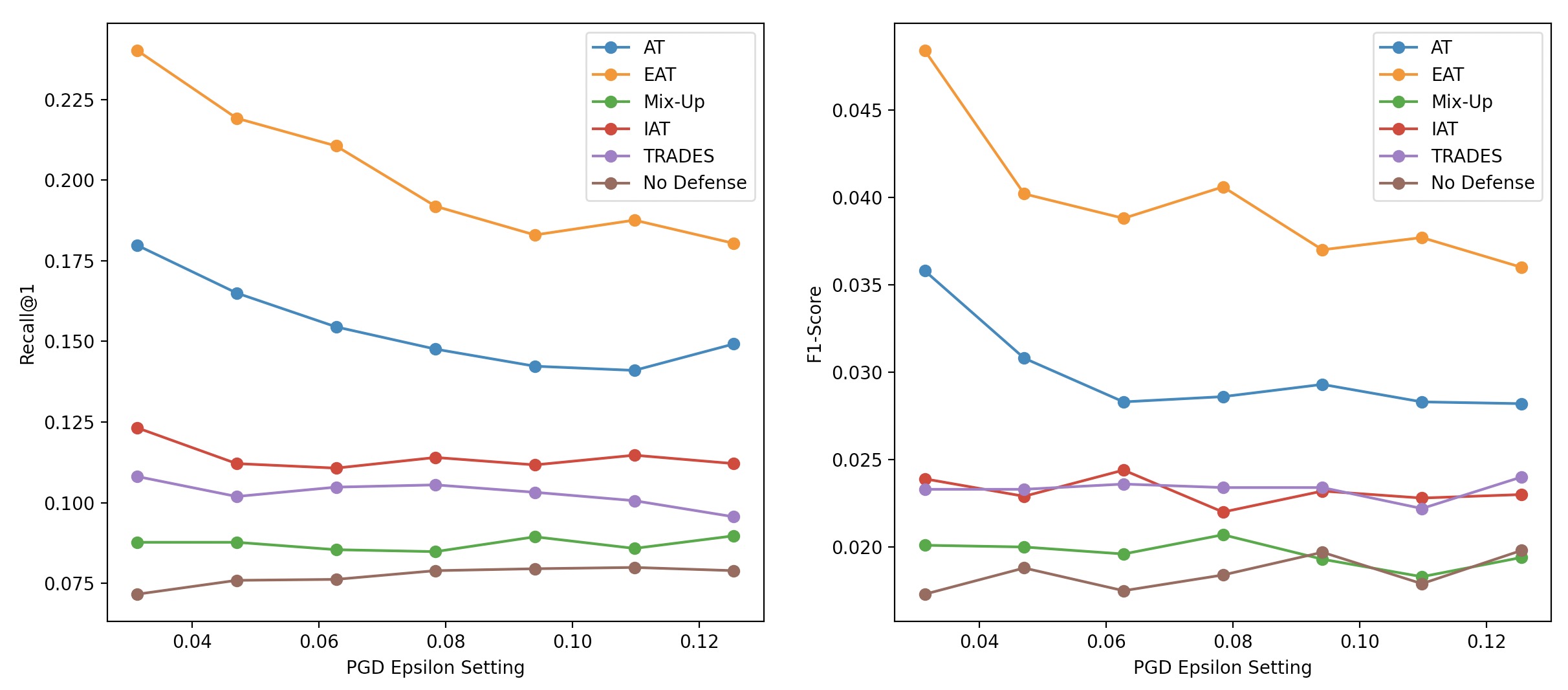}
\end{center}
  \caption{Recall@1 and F1-Score for different $\varepsilon$ settings of the PGD attack with the different defenses on CUB200 (Training with the MobileNetV2)}
\label{fig:eps-recall@1-f1-mobi-cub200}
\end{figure}

\textbf{The robustness under different attack budgets $\varepsilon$ settings.} We evaluate our EAT defense in comparison with the other defenses under different $\varepsilon$ settings (8/255, 12/255, 16/255, 20/255, 24/255, 28/255, and 32/255). As is shown in Figure \ref{fig:eps-recall@1-f1-mobi-cub200}, the EAT defense still outperforms other defenses. The F1-Score and the Recall@1 decrease when increasing the $\varepsilon$ of the PGD attack. There are also some fluctuations due to the randomness of the iteration step. 



\begin{figure}[]
\begin{center}
\includegraphics[width=\linewidth]{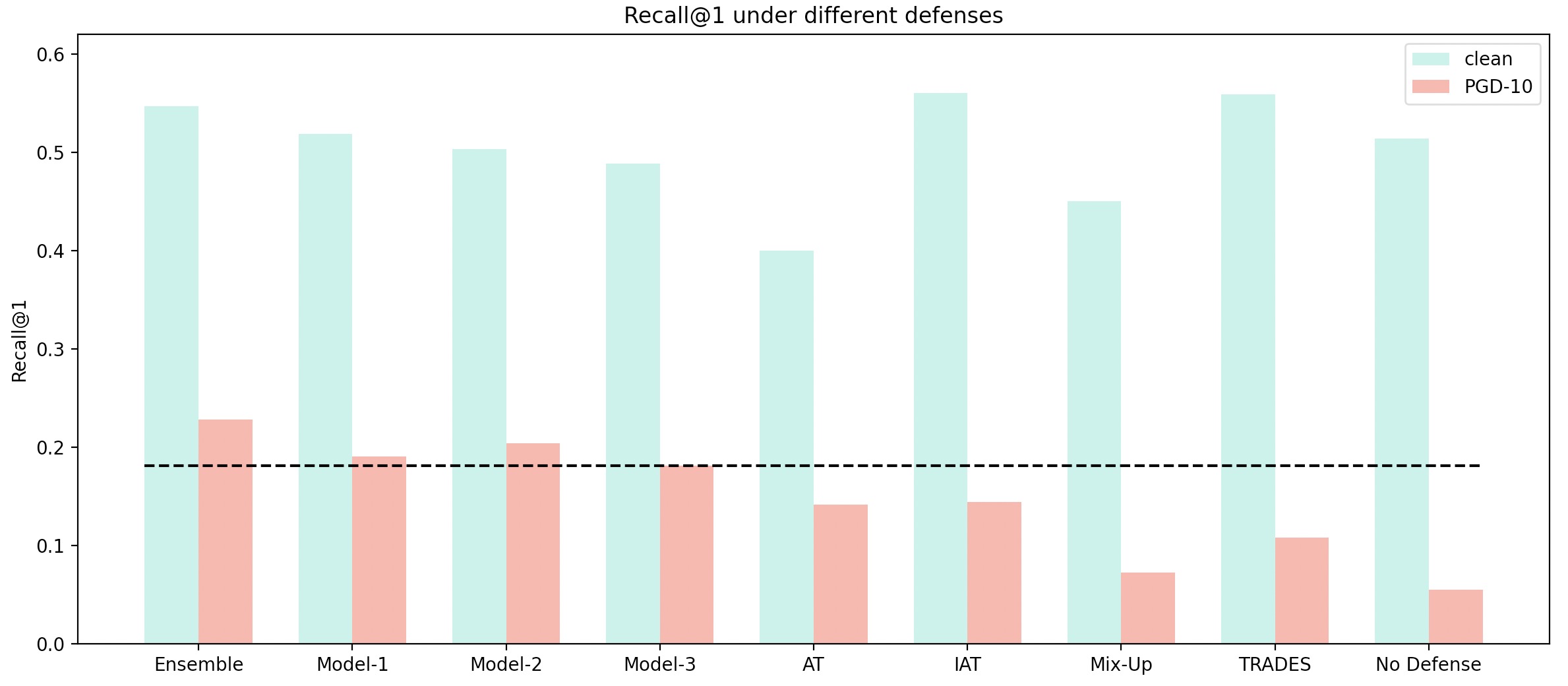}
\end{center}
  \caption{Recall@1 for the individual model of the EAT defense in comparison with other defenses on CUB200 (Training with MobileNetV2)}
\label{fig:recall@1-f1-ind-model-cub200}
\end{figure}

\textbf{The adversarial robustness of the individual model in the whole ensemble}. As is shown in Figure \ref{fig:recall@1-f1-ind-model-cub200}, every individual model of the ensemble is still more robust than other defenses and it is better to construct the ensemble and apply our voting mechanism for achieving higher robustness.



\subsection{Ablation Experiments}

\begin{table}[]
\begin{adjustbox}{width=\columnwidth,center}
\begin{tabular}{cccccc}
\hline
Method                              &        & NMI            & F1-Score      & Recall@1       & Recall@8       \\ \hline
\multirow{2}{*}{Naive Ensemble}     & Clean  & 58.95          & 20.93         & 53.47          & 81.76          \\
                                    & PGD-10 & 34.76          & 3.26          & 17.91          & 40.91          \\ \hline
\multirow{2}{*}{EAT w/o Data Split} & Clean  & 59.28          & 21.25         & 54.49          & 82.02          \\
                                    & PGD-10 & 36.13          & 4.23          & 21.46          & 45.25          \\ \hline
\multirow{2}{*}{EAT w/ Data Split}  & Clean  & 56.91          & 19.28         & 52.78          & 80.48          \\
                                    & PGD-10 & \textbf{38.02} & \textbf{4.84} & \textbf{24.02} & \textbf{47.49} \\ \hline
\end{tabular}
\end{adjustbox}
\caption{The ablation experiment on CUB200 with MobileNetV2 under for presenting the benefit of the split mechanism.}
\label{tab:mob-cub200-split}
\end{table}

\begin{table}[ht]
\begin{center}
\begin{adjustbox}{width=\columnwidth}
\begin{tabular}{ccccccccc}
\hline
Method                                 &        & NMI            & F1-Score       & Recall@1       & Recall@10      & Recall@20      \\ \hline
\multirow{2}{*}{\textbf{PAL}}          & Clean  & 86.88          & 14.95          & 60.97          & 82.69          & 86.70          \\ \cline{2-7} 
                                       & PGD-10 & 82.47          & 3.36           & 16.25          & 32.86          & 39.05          \\ \hline
\multirow{2}{*}{\textbf{PAL + AT}}     & Clean  & 86.71          & 14.19          & 57.29          & 80.41          & 84.79          \\ \cline{2-7} 
                                       & PGD-10 & 84.68          & 8.51           & 38.06          & 62.46          & 69.20          \\ \hline
\multirow{2}{*}{\textbf{PAL + Mix-Up}} & Clean  & 86.88          & 14.85          & 60.77          & 82.75          & 86.68          \\ \cline{2-7} 
                                       & PGD-10 & 82.46          & 3.43           & 16.66          & 33.44          & 39.53          \\ \hline
\multirow{2}{*}{\textbf{PAL + IAT}}    & Clean  & 86.76          & 14.37          & 57.98          & 80.86          & 85.54          \\ \cline{2-7} 
                                       & PGD-10 & 84.78          & 8.80           & 38.27          & 63.03          & 69.72          \\ \hline
\multirow{2}{*}{\textbf{PAL + TRADES}}   & Clean  & 87.35          & 16.18          & 63.37          & 84.61          & 88.37          \\ \cline{2-7} 
                                       & PGD-10 & 81.95          & 2.45           & 11.36          & 24.38          & 28.99          \\ \hline
\multirow{2}{*}{\textbf{PAL + EAT}}    & Clean  & 86.82          & 14.40          & 58.69          & 79.58          & 84.34          \\ \cline{2-7} 
                                       & PGD-10 & \textbf{85.30} & \textbf{10.03} & \textbf{43.16} & \textbf{66.16} & \textbf{72.64} \\ \hline
\end{tabular}
\end{adjustbox}
\end{center}
\caption{The comparison of five defenses on In-Shop with MobileNetV2 under the white-box adversarial example attack.}
\label{tab:mob-in-shop}
\end{table}

\textbf{The comparison of the Naive Ensemble Method}
As is shown in Table \ref{tab:mob-cub200-split}, we perform the naive ensemble method (i.e. Training three models individually and construct a simple ensemble through the voting mechanism) on CUB200. Our EAT method presents better robustness than the naive ensemble method.

\textbf{The robustness gain of the split mechanism}
As is shown in the Table \ref{tab:mob-cub200-split}, we also analyze the gain of our split mechanism. Our split mechanism brings the 9.13\% robustness improvement on CUB200.
\section{Conclusion}

In this work, we analyze the adversarial robustness of  DML models when they are used in clustering scenarios. We find that the simple adaptions of defenses designed for deep classification models, including AT, Mix-Up, IAT, and TRADES, cannot work well for DML models.
To defend against adversarial examples, we propose a new method, the Ensemble Adversarial Training (EAT), which takes advantage of ensemble learning and adversarial training.  EAT promotes the diversity of the ensemble to make the adversarial training more powerful by exploiting dataset arrangement, and employs a self-transferring mechanism to further improve the adversarial robustness of the ensemble. We implement and evaluate our EAT defense on three popular datasets with two commonly-used datasets. The experimental results show that EAT greatly outperforms defenses that are simply adapted from defenses designed for deep classification models.


\bibliography{egbib}

\end{document}